\pdfoutput=1

\documentclass[11pt]{article}

\usepackage[preprint]{acl}

\usepackage{times}
\usepackage{booktabs}
\usepackage{latexsym}
\usepackage{graphicx}
\usepackage[T1]{fontenc}

\usepackage[utf8]{inputenc}

\usepackage{microtype}

\usepackage{inconsolata}

\usepackage{graphicx}

\usepackage{float}
\restylefloat{table}

\usepackage{array}
\usepackage{multirow}
\usepackage{multicol}

\usepackage{todonotes}

\title{Fact-checking AI-generated news reports: Can LLMs catch their own lies?}

\author{Jiayi Yao \\
  Brandeis University \\
  \texttt{jiayiyao@brandeis.edu} \\\And
  Haibo Sun \\
  Brandeis University \\
  \texttt{hsun@brandeis.edu} \\ \And
Nianwen Xue \\
  Brandeis University \\
  \texttt{xuen@brandeis.edu} \\}

\begin{document}
\maketitle
\begin{abstract}
In this paper, we evaluate the ability of Large Language Models (LLMs) to assess the veracity of claims in ``news reports'' generated by themselves or other LLMs. Our goal is to determine whether LLMs can effectively fact-check their own content, using methods similar to those used to verify claims made by humans. 
Our findings indicate that LLMs are more effective at assessing claims in national or international news stories than in local news stories, better at evaluating static information than dynamic information, and better at verifying true claims compared to false ones. We hypothesize that this disparity arises because the former types of claims are better represented in the training data.
Additionally, we find that incorporating retrieved results from a search engine in a Retrieval-Augmented Generation (RAG) setting significantly reduces the number of claims an LLM cannot assess. However, this approach also increases the occurrence of incorrect assessments, partly due to irrelevant or low-quality search results. This diagnostic study highlights the need for future research on fact-checking machine-generated reports to prioritize improving the precision and relevance of retrieved information to better support fact-checking efforts. Furthermore, claims about dynamic events and local news may require human-in-the-loop fact-checking systems to ensure accuracy and reliability.

\end{abstract}

\section{Introduction}

Large Language Models (LLMs) have revolutionized the field of Natural Language Processing (NLP), effortlessly performing tasks that were traditionally considered highly challenging. Their performance is particularly impressive in generating natural language text. Models like GPT-4 can generate coherent, fluent summaries, accurately translate text between languages (especially those with a strong online presence and ample training data), and refine human writing to enhance fluency and appropriateness in tone and style for specific purposes. This technology has the potential to significantly increase productivity across many industries, offering endless applications. However,
with this potential also come risks if they are not used properly. One of the main risks is that they can be easily used to generate convincing and yet factually incorrect text, either intentionally or unintentionally.  For example, with a simple prompt like ``Generate a news report about volcano eruption in Massachusetts, USA'', GPT-4 can generate a news report starting with this first paragraph:

\begin{quote}
''Massachusetts, USA – May 29, 2024 – In an unprecedented and shocking event, a volcanic eruption has occurred in the state of Massachusetts, an area not typically associated with volcanic activity. The eruption took place early this morning in the central part of the state, near the town of Worcester, sending residents and scientists alike into a state of disbelief and concern.'' 
\end{quote}

Although there has never been a volcanic eruption in reality, the news report is coherent and fluent. 
Coupled with modern media platforms, such LLM-generated content can quickly spread and reach a large audience. An example is the emergence of AI ``news'' farms that produce news reports with LLMs to generate advertising revenue with little concern for their impact on society \cite{puccetti2024ainewscontentfarms}. The machine-generated reports can cause confusion and chaos and disrupt the proper functioning of the society.  In fact, studies show that false news tends to spread ``farther, faster, deeper'' than true news, as it often contains novel content that people are more likely to share \cite{Vosoughi2018science}. 

In this study, we present experimental results
to answer the questions of whether LLMs are capable of telling if the news stories they generate are truthful and how well they can catch factually incorrect claims in those news stories. 
We generated 92 news stories with a simple prompt such as ''Write a story about Kobe Bryant rejoining the Lakers'' with two LLMs, GPT-4o \cite{openai2024gpt4technicalreport} and GLM \cite{du-etal-2022-glm}, all stories with some incorrect claims. These false stories vary by how untruthful they are. Some stories report events that are simply impossible, such as a story about Kobe Byrant rejoining the Lakers, as the former Lakers star has passed away. Some stories report events that are not out of the realm of possibilities but are highly unlikely, such as a volcano eruption in Massachusetts, as the area is not known for volcanic activities. Other stories are about events that have actually happened or are scheduled to happen, but with the wrong time, location, or participants.  

We performed our experiments in two settings. In the first setting, we simply provided the full story to GPT-4o and GLM as input and asked if they are truthful.
In the second setting, we manually decompose each story into individual checkable \textit{atomic} claims. A checkable claim can be either an event with specific participants, location, or time, or they can be a state  (e.g., Massachusetts borders New Hampshire) or recurring event that holds for an extended period of time such that the exact time is irrelevant. We perform manual ``decontexutalization'' \cite{choi-etal-2021-decontextualization} on these checkable claims so that they can be verified outside of the context of their document. For this setting, we also experimented with using these checkable claims as queries, providing the results retrieved via the Google Search Serper Api \footnote{\hyperref[Serper Api]{\url{https://serper.dev/}}} to GPT-4 to assist in evaluating the veracity of the claims within a Retrieval-Augmented Generation (RAG) framework \cite{Lewis2020rag}.

The results of our experiments 
show that GPT-4o and GLM are very good at detecting stories that contain incorrect claims (and all of them do) when they involve well-known entities (e.g., Kobe Bryant rejoining the Lakers), but they are quite uncertain about recent events that are unlikely. 
At the level of atomic claims,  a significant proportion  of them  are incorrectly assessed: either a factually correct claim is judged to be wrong or a factually wrong claim is identified as being correct. For an even larger proportion of atomic claims, the LLMs simply cannot decide.
When provided with results retrieved via the Google Search Serper Api, the number of non-assessments decreases significantly, accompanied by an increase in both correct and incorrect assessments. Interestingly, even when the Google Search Serper Api returns no results for a claim, GPT-4 still attempts to provide an assessment instead of declining to answer. It appears that simply knowing no results were retrieved is enough to prompt GPT-4 to make a guess. Even with RAG, there is still a significant proportion of claims that the LLM cannot provide an assessment.
This means that any solution to fact-checking machine-generated news reports needs to include functionalities on checking claims about new event occurrences that are not checkable against existing knowledge sources. While there has been recent research that shows the promise of using external resources or tools to improve the factuality of LLMs \cite{Gou2023CRITICLL}, such an approach is not applicable to fact-checking machine-generate news stories and novel human-in-the-loop methods may need to be developed to check such claims.

The rest of the paper is organized as follows. In Section \ref{sec:related}, we discuss related work. In Section \ref{sec:method}, we present our method for generating news stories, extract ``atomic'' claims, using LLMs to assess the veracity of these stories and claims, and manually verifying the assessments performed by LLMs themselves. We present experimental results in Section \ref{sec:experiment}, and discuss these results in Section \ref{sec:discussion}. We conclude in Section \ref{sec:conclusion}.

\section{Related Work}
\label{sec:related}

\paragraph{Dataset statistics and comparison.}
Our data set is most similar to FactScore \cite{min-etal-2023-factscore} in that both consist of long-form texts generated by LLMs; however, two key differences separate our data sets. First, while FactScore focuses on biographies of Wikipedia entities, our dataset consists of LLM-generated news reports that include time-sensitive content, making them inherently harder to fact check due to the absence of a preexisting knowledge source. Second, while every short sentence in the biographies of FactScore is treated as an independent factual claim, our news reports often contain vague or subjective content, necessitating manual extraction of only those claims that are verifiably checkable.
The following data sets are also broadly related to ours, but there are significant differences.  Datasets like PROPANEWS (arXiv:2203.05386)  are created by replacing sentences in real news articles with plausible but fake content to mimic factual claims made by humans.  The FEVER data set consists of individual claims verified by Wikipedia. The EX-FEVER data set is also based on Wikipedia but requires multi-hop reasoning to fact-check to enhance explainability. The AVeriTeC data set contains real-world claims that can be checked against web sources. All these data sets are collections of individual claims created with the assumption that there is a knowledge source against which these claims can be verified.

\paragraph{Fact-checking human or machine-generated content.} There is an active NLP research community focused on developing automatic methods to fact-check false claims, such as those made by politicians \cite{nakov2021ijcai,deng-etal-2024-document,yuan-vlachos-2024-zero,schlichtkrull-etal-2024-automated}. 
There is also more recent work on fact-checking machine-generated content \cite{min-etal-2023-factscore, wang2024factcheck, fadeeva-etal-2024-fact}. Previous work on fact-checking false claims made by either humans or machines typically assumes there is an information source, usually a published source on the Internet, against which the claims can be checked. However, events reported in machine-generated news stories that we are interested in, such as the volcano eruption example, are often assumed to be new occurrences that cannot be cross-verified against any existing public sources, although they may still contain claims about the real world that can be fact-checked. This poses novel challenges that are not present in people biographies used in previous studies \cite{min-etal-2023-factscore,fadeeva-etal-2024-fact}.

\section{Method}
\label{sec:method}
Our experiment on fact-checking LLM-generated news stories consists of four steps. First, we use two LLMs to generate a set of news stories with varying levels of factual inaccuracy. Next, from these stories, we manually extract verifiable atomic claims and decontextualize them, creating standalone claims that can be verified independently of the original story. In the third step, we prompt each LLM to evaluate the veracity of news stories generated by itself or the other LLM, as well as to assess the individual atomic claims. Finally, we conduct a human evaluation to determine the accuracy of the LLMs’  veracity assessments.

\subsection{News Report Generation with LLMs}
To evaluate the claim verification capabilities of GPT and GLM, we first prompt both models to generate a set of 92 news articles, including 47 news artiles generated by GPT-4o and 44 articles generated by GLM. Each prompt is designed around scenario-based inputs that intentionally contain factual inconsistencies.  
The following is an example prompt that contains a time error, as the time of 2024 Australian Open women's final is January 27, not January 20:
\begin{quote}
"Generate a news report about Aryna Sabalenka winning the 2024 Australian Open Women's final, held at Rod Laver Arena on \underline{January 20}, as Aryna Sabalenka beat Zheng Qinwen (6-3, 6-2)." 
\end{quote}

All these inconsistencies are designed around four critical aspects of a scenario: the event itself, along with its time, location, and participants. 
To rigorously test the models’ understanding of both nationally recognized and locally relevant information, we control the scope of the generated content by introducing both local and national news categories. The distinction between these categories serves as a critical factor in our evaluation, allowing us to evaluate how effectively each model handles claims involving specific local information versus those based on widely known national knowledge. This is motivated by prior research suggesting that LLMs may have greater exposure to widely discussed national or international events, given the nature of the large, diverse datasets they are trained on \cite{kandpal2023largelanguagemodelsstruggle}. 
When generating the news stories, we ensure that the same general template is used for all prompts, varying only the scenarios for each different story.
By using consistent prompts, we ensure that differences in model performance can be attributed to the model's capabilities rather than variability in the inputs. This approach allows us to build a diverse and representative dataset that rigorously tests each LLM’s ability to identify and evaluate issues across different aspects of the generated content.

\subsection{Manual claim Extraction}
After generating the news reports, we manually extracted all checkable claims from the GPT-generated content. Each claim is a clear, verifiable statement with specific details such as time, location, participants, or events. We adhered to criteria that required each checkable claim to contain precise, unambiguous information—such as exact dates, locations, or identifiable participants. Vague or generic statements, like ``Sabalenka had a great match'' were excluded, as they lack objective, verifiable details. This approach ensured that only claims containing concrete, factual information were selected for manual extraction. We manually decontextualize claims by resolving pronominal and other anaphoric expressions, and by supplementing events with time, location, and participant details when they are clear from the context, ensuring that each claim is independently verifiable.

The following are example claims illustrating various types of factual inaccuracies:

\begin{itemize}
\setlength{\itemsep}{-3pt}
\item  \textbf{Time error:} “Aryna Sabalenka triumphed over Zheng Qinwen to win the 2024 Australian Open Women's final at Rod Laver Arena on \underline{January 20, 2024.}”

\item \textbf{Location error:} “Aryna Sabalenka played against Zheng Qinwen in the 2024 Australian Open Women's final at \underline{Margaret Court Arena} on January 27.”

\item \textbf{Event error:} “In the third set of the 2024 Australian Open Women's Final at Rod Laver Arena on January 27, \underline{Zheng Qinwen broke Aryna Sabalenka's serve} \underline{at 5-5 and won the set 7-5 to clinch the} \underline{championship}.”

\item \textbf{Participant and location error:} “\underline{Naomi Osaka and Iga Swiatek} are battling for the prestigious Grand Slam title at the 2024 Australian Open Women's Final at \underline{the Margaret Court Arena} on January 27, 2024.”
\end{itemize}

Each article typically yields between 10-20 checkable decontextualized claims, depending on its length and complexity. 
This process ensures that the claims include all the necessary contextual information required for verification, maintaining the integrity and relevance of the claims within the broader context of the news reports. 
From the 92 articles we have extracted 1,337 total atomic claims, including 697 claims from the 47 news reports generated by GPT-4o, and 640 claims from the 44 reports generated by GLM.  The breakdown of the error types by entire articles and atomic claims is presented in Table \ref{tab:error-breakdown}. Note that some articles or claims may contain multiple types of errors.

\begin{table}[ht]
\centering
\resizebox{\columnwidth}{!}{%
\begin{tabular}{lcccc}
\toprule
\textbf{Generator} & \multicolumn{2}{c}{\textbf{GPT-4o}} & \multicolumn{2}{c}{\textbf{GLM}} \\ 
\cmidrule(lr){2-3} \cmidrule(lr){4-5}
\textbf{Error type} & \textbf{Whole articles} & \textbf{Atomic claims} & \textbf{Whole articles} & \textbf{Atomic Claims}  \\
\midrule
Event        & 30   & 283 & 28 & 272     \\
participant  & 13   & 69  & 12 & 51   \\
time         & 3    & 40  & 3  & 47    \\
location     & 17   & 166 & 17 & 191    \\
\bottomrule
\end{tabular}%
}
\caption{\label{tab:error-breakdown}Count of error types in entire articles and atomic claims }
\end{table}

\subsection{Claim verification with LLMs}
Both GPT-4o and GLM models are tasked with verifying the veracity of each entire article as well as each atomic claim. To assess claim veracity, we prompted GPT-4o and GLM to evaluate the accuracy of all 92 news articles and their corresponding atomic claims. The following are the prompts we use for the evaluation: 
\begin{itemize}
\setlength{\itemsep}{-3pt}
    \item 
\textbf{Article-level prompt:} “Today is August 1st, 2024. You are a helpful assistant that performs the below tasks: verify if the following news is accurate or false. Respond as concisely as possible.”

\item \textbf{Claim-level prompt:} “Today is August 1st, 2024. You are a helpful assistant that performs the below tasks: verify if the following claim extracted from a news report is accurate or false. Respond as concisely as possible.”
\end{itemize}

The models are first prompted to assess the veracity of each entire article and provide a rationale for their evaluations. They are then prompted to evaluate the veracity of each atomic claim extracted from the articles, along with a rationale for each assessment. Three different prompting approaches are used in this pipeline.

\subsubsection{Deterministic Prompting (Temperature 0.0)} 
We prompt the models to provide a singular, deterministic evaluation for each article or claim. Setting temperature to 0 minimizes randomness and allows us to observe the models' baseline claim verification performance under controlled conditions.

\subsubsection{Self-consistency Prompting (Temperature 1.0)} 
We use a higher temperature setting (1.0) to introduce variability in the responses of the models. Models are prompted multiple times (5 times per article / claim in our experiment), and a majority voting mechanism is used to determine the final assessment. This setting simulates the potential variability in model reasoning and robustness across multiple prompts.

In each instance, the model outputs a determination (correct or false) along with a rationale for its assessment. These rationales are crucial for error analysis, offering insights into whether the model's reasoning aligns with the factual basis of the claim.

\subsubsection{RAG Prompting} We queried the Google Search Serper Api with manually extracted atomic claims and incorporated the retrieved results into the prompt for GPT-4 when evaluating the veracity of claims within a Retrieval-Augmented Generation (RAG) framework. The goal of this experiment was to assess whether providing search results improves the evaluation accuracy of LLMs. Due to cost constraints and the length limitation of the search engine, we did not perform this experiment with the entire article. Instead, we focused on atomic claims extracted from news reports generated by GPT-4 itself, assuming the results would generalize to other settings.

\subsection{Comparing model verification with human judgments}
To validate the models' evaluations, we manually verify each claim by conducting targeted web searches and cross-referencing the findings with our existing information. We use independent online sources, including reputable news databases, fact-checking websites, and government records. The human judgments serve as the gold standard for evaluating model assessments, enabling us to quantify both false positives and false negatives in the models' evaluations.
Additionally, we performed error analysis to understand whether the type of news (local vs. national) and the type of claim (states vs events, true vs false claims) had a measurable impact on the model’s performance. Special attention was paid to cases where the models provided no assessment, incorrect reasoning, or inaccurate evaluations.

\section{Experiments}
\label{sec:experiment}
We conduct a comprehensive set of experiments to evaluate the performance of GPT-4o and GLM models in verifying claims within generated news articles. Both models are assessed in the contexts of local and national news generation, with claim verification performed across all relevant dimensions. For the claim verification task, we classify the assessment results into five possible categories, as outlined below:

\begin{itemize}
    \item 
    \setlength{\itemsep}{-5pt}
\textbf{Correct Assessment (CA):} The model correctly identifies the veracity of the claim without providing a rationale.
\item \textbf{Correct Assessment and Correct Reasoning (CA/CR):} The model correctly identifies the veracity of the claim and provides a correct justification for its assessment.
\item \textbf{Correct Assessment and Wrong Reasoning (CA/WR):} The model correctly classifies the claim but with flawed reasoning.
\item \textbf{Wrong Assessment (WA):} The model incorrectly classifies the veracity of the claim.
\item \textbf{No Assessment (NA):} The model fails to provide any assessment.
\end{itemize}

For examples of each type of assessment, please see Appendix \ref{appendix:assessment-type}.

\begin{table}[ht]
\centering
\resizebox{\columnwidth}{!}{%
\begin{tabular}{lcccccc}
\toprule
\textbf{Generator} & \multicolumn{3}{c}{\textbf{GPT-4o}} & \multicolumn{3}{c}{\textbf{GLM}} \\ 
\cmidrule(lr){2-4} \cmidrule(lr){5-7}
\textbf{Evaluator} & \textbf{GPT-4o} & \textbf{GPT-4-turbo} & \textbf{GLM-4} & \textbf{GPT-4o} & \textbf{GPT-4-turbo} & \textbf{GLM-4} \\
\midrule
CA        & 1    & 1   & 0 & 1    & 0 & 0 \\
CA/CR     & 30   & 26  & 37 & 31  & 31 & 31 \\
CA/WR     & 5    & 2   & 1 & 5    & 6  & 2 \\
WA        & 9    & 8   & 8 & 2    & 2  & 10 \\
NA        & 2    & 10  & 1 & 6    & 6  & 2 \\
\midrule
Total     & 47   & 47  & 47 & 45   & 45 & 45 \\
\bottomrule
\end{tabular}%
}
\caption{\label{tab:entire-report}Count of LLM-generated articles for each assessment category}
\end{table}

\subsection{Entire news articles}
Table \ref{tab:entire-report} presents the performance data of GPT-4 (gpt-4o-20240806 and gpt-4-turbo-20240409) and GLM-4 (GLM-4-0520) in evaluating entire articles. Both models were prompted to generate news reports, followed by self-evaluation and cross-evaluation of the generated articles.

GPT-4 and GLM-4 demonstrate comparable performance in the number of correct and incorrect assessments they produce. In contrast, GPT-4 Turbo is more likely to refrain from making assessments, reflecting a more cautious approach compared to GPT-4o and GLM-4. This suggests that GPT-4-turbo prioritizes minimizing errors, even if it results in fewer overall judgments.

\subsection{Individual atomic claims}

In evaluating LLMs in verifying atomic claims, we conducted experiments with GPT-4o and GLM-4 to ensure our findings are generalizable across LLMs. The performance of GPT and GLM models was assessed across different temperature settings to better assess their strengths and limitations in claim verification tasks. Both models were tasked with verifying the veracity of claims extracted from LLM-generated news articles, with their evaluations measured using the identical 5-dimensional protocol we use for entire articles. 

The assessment results are presented in Table \ref{tab:claims} and we can make severval key observations. First, GPT-4o consistently provides more correct assessments (including those with and without correct reasoning) than GLM, regardless of whether it is evaluating claims from articles it generated or those generated by GLM. This trend holds across all temperature settings. Interestingly, both GPT-4o and GLM produce more incorrect assessments (WA) when evaluating claims from articles they generated themselves. The most notable finding is the high number of cases with no assessment (NA), with GLM showing a significantly higher number (about 20\%) of no assessments than GPT-4. 

\begin{table*}[ht]
\centering
\resizebox{\textwidth}{!}{%
\begin{tabular}{lllllllll}
\toprule
\textbf{Generator} & \multicolumn{4}{c}{\textbf{GPT-4o}} & \multicolumn{4}{c}{\textbf{GLM}} \\ 
\cmidrule(lr){2-5} \cmidrule(lr){6-9}
\textbf{Evaluator} & \textbf{GPT/0} & \textbf{GPT/1} & \textbf{GLM/0} & \textbf{GLM/1} & \textbf{GPT/0} & \textbf{GPT/1} & \textbf{GLM/0} & \textbf{GLM/1} \\ \hline
CA (\%)  &38(5.45)  &44(6.31)&1(0.14)&0(0.00)&15(2.34)&14(2.19)&3(0.47)&5(0.78)  \\
CA/CR(\%)&306(43.90)&291(41.75)  & 271(38.88) & 276(39.60) & 349(54.53)  & 353(55.16)  & 312(48.75) & 306(47.81) \\
CA/WR(\%)&5(0.72)   & 10(1.43)   & 13(1.87)  & 15(2.15)  & 42(6.56)   & 31(4.84)   & 24(3.75)  & 24(3.75)  \\
WA(\%)   &33(4.73)  &42(6.03)   & 12(1.72)  & 14(2.00)  & 9(1.40)    & 12(1.88)   & 15(2.34)  & 29(4.53)  \\
NA(\%)   &315(45.19)&310(44.48)  & 400(57.39) & 392(56.24) & 225 (35.16) & 230(35.94)  & 286(44.69) & 276(43.13) \\
\midrule
Total    & 697  & 697  & 697 & 697 & 640  & 640  & 640 & 640 \\
\bottomrule
\end{tabular}%
}
\caption{\label{tab:claims} Count and percentage of individual atomic claims for each assessment category across models at different temperature settings. GPT/0 and GPT/1 indicate GPT at temperature 0 and 1 respectively. Similarly, GLM/0 and GLM/1 indicate GLM at temperature 0 and 1.}
\end{table*}

\subsubsection{Claims in National vs Local news stories}

We also attempted to evaluate the ability of LLMs to assess claims in national and local news stories. The following are example claims from national and local news stories we generated with LLMs:
\begin{itemize}
\setlength{\itemsep}{-3pt}
\item  \textbf{Claims in local news:} The free rave hosted by Watertown, MA on July 15, 2024 will be held at Arsenal Park.
\item  \textbf{Claims in national or international news:} The 2024 Paris Olympics opening ceremony is set to take place on July 26.
\end{itemize}

\begin{table*}[h!]
\centering
\resizebox{\textwidth}{!}{%
\begin{tabular}{lllllllllll}
\toprule
\textbf{Generator}
& \multicolumn{5}{c}{\textbf{GPT-4o}}
& \multicolumn{5}{c}{\textbf{GLM}} \\
\cmidrule(lr){2-6} \cmidrule(lr){7-11}
\textbf{Evaluator} 
& \textbf{Subt.}
& \textbf{GPT/0}
& \textbf{GPT/1}
& \textbf{GLM/0}
& \textbf{GLM/1}
& \textbf{Subt.}
& \textbf{GPT/0} 
& \textbf{GPT/1} 
& \textbf{GLM/0} 
& \textbf{GLM/1} \\
\midrule
National(\%) & 496 & 193(38.91) & 208(41.94) & 252(50.81) & 247(49.80) & 462 & 143(30.95) & 141(30.52) & 194(41.99) & 197(42.64) \\
Local(\%) & 201   & 160(79.60) & 154(76.62) & 173(86.07) & 174(86.57) & 178 & 133(74.72) & 132(74.16) & 131(73.60) & 132(74.16) \\
\midrule
Total  & 697 & 353(50.65) & 362(51.94) & 425(60.98) & 421(60.40) & 640 & 276(43.13) & 273(42.66) & 325(50.78) & 329(51.41) \\
\bottomrule
\end{tabular}%
}
\caption{\label{tab:national-local} Errors from evaluating claims in national or local news. Each cell represents the percentage of claims that are incorrectly assessed for that category (national vs local), with the last row representing the number of errors / the total claims for that generator.}
\end{table*}

Table \ref{tab:national-local} 
presents a comparative error analysis of GPT and GLM models when evaluating claims from national and local news sources, across different temperature settings. Errors in assessments include cases where the model provides the correct assessment with wrong reasoning (CA/WR), wrong assessment (WA), or no assessment (NA). As we can see from the table, while GPT slightly outperforms GLM as indicated by the generally lower number of errors, the error rate is relatively consistent across temperatures.
The most notable finding is the substantial difference in error rates between the models' assessments of claims from national and local news, with significantly higher error rates for local news than national news. 

One possible explanation is that claims in national news often pertain to major events or widely recognized topics that are well-documented across diverse online sources, making these claims more likely to appear in the models' training data and thus easier to assess. In contrast, claims in local news may involve niche, region-specific issues that receive limited attention and documentation, leaving the models less prepared to verify such claims accurately. This discrepancy highlights how the scope and distribution of training data can impact the models’ performance in evaluating claims with different degrees of specificity and familiarity.

\subsubsection{Assessment of true claims vs false claims}

Table \ref{tab:combined} evaluates the accuracy of LLMs in assessing both factually correct and wrong claims. We analyze whether the LLMs make accurate or inaccurate assessments when presented with claims that are either true or false. Correct Assessment includes cases where 
(i) the claim is factually true, and the LLM assesses it as true.
(ii) The claim is factually false, and the LLM assesses it as false. 
And wrong assessment includes cases where 
(i) the claim is factually false, but the LLM assesses it as true and (ii) the claim is factually true, but the LLM assesses it as false.

We aim to investigate whether there is a difference in the accuracy with which LLMs assess factually true versus false claims. Our hypothesis is that factually true claims are more likely to be represented in the training data than factually false ones, making it more probable that factually false claims will be incorrectly assessed. Our hypotheis is born out, as results in Table \ref{tab:combined} show that  both the GPT and GLM generally have a higher rate of correct assessments when the claim was factually correct while both models struggle with factually wrong claims and made wrong assessments. Among all the cases where the model made correct assessments but provided incorrect reasoning,  a considerable portion of them is from claims that are factually wrong. This suggests that while the model can arrive at the correct conclusion, its internal logic or justifications may still be flawed, which happens mostly when the claims are factually incorrect.

\begin{table*}[h!]
\centering
\resizebox{\textwidth}{!}{%
\begin{tabular}{lllllllll}
\hline
\multicolumn{1}{l}{\textbf{Generator}} & \multicolumn{4}{c}{\textbf{GPT-4}} & \multicolumn{4}{c}{\textbf{GLM}} \\
\cmidrule(lr){2-5} \cmidrule(lr){6-9}
\multicolumn{1}{l}{\textbf{Evaluator}} & \multicolumn{2}{c}{\textbf{GPT/0}} & \multicolumn{2}{c}{\textbf{GLM/0}} & \multicolumn{2}{c}{\textbf{GPT/0}} & \multicolumn{2}{c}{\textbf{GLM/0}}   \\ \cmidrule(lr){2-5} \cmidrule(lr){6-9}
\textbf{Veracity}                     & FC(\%) & FW(\%)& FC(\%) & FW(\%) &FC(\%) &FW(\%) &FC(\%) &FW(\%) \\ \hline
CA (CR) & 143 (87) & 201 (38)& 135 (82)& 137 (26)  & 92 (89) & 272 (51) & 86 (84) & 229 (43) \\ 
CAWR   & 0 (0.0) & 5 (0.9) & 1 (0.6) & 12 (2.3) & 2 (1.9) & 40 (7.4) &0 (0.0)& 24 (4.5) \\ 
WA     & 4 (2.4) & 29 (5.5) & 4 (2.4) & 8 (1.5) & 1 (1.0) & 8 (7.8) & 5 (4.9)& 10 (1.9) \\ 
NA     & 18 (11)&297 (56)&25 (15)&375 (71)&8 (7.8)&217 (40)&12 (12)&274 (51)\\ \hline
Total& 165&532&165&532&103&537&103&537\\
\hline
\end{tabular}%
}
\caption{\label{tab:combined} Comparison of LLM assessment accuracy for factually correct (FC) and factually incorrect (FW) claims with GPT and GLM as evaluators at 0 temperature.}
\end{table*}

\subsubsection{State and event claims}
\label{state-event}
We also experimented with asking LLMs to assess claims that are linguistic states and those that are not. Here, a state refers to a specific condition or phase in the existence of something, characterized by stability and consistency over time, whereas a non-state claim typically involves an event, signifying a significant occurrence that brings about change. A non-state claim is typically associated with a time, location, and participants.  The following shows example claims categorized as state and non-state:
\begin{itemize}
\setlength{\itemsep}{-3pt}
\item  \textbf{State claim:} Aryna Sabalenka is Belarusian.
\item  \textbf{Non-state claim:} The 2024 Australian Open Women's final was held at Margaret Court Arena on January 27.
\end{itemize}

We hypothesize that LLMs perform better on state claims because states are more stable and likely to be documented in training data, whereas events are often new and undocumented. Consequently, LLMs are more prone to errors, including wrong assessments (WA) and no assessments (NA), when evaluating non-state claims, as supported by the higher error rates observed for these claims. This hypothis is largely born out by the higher error rate for non-states than states. We also observed a significant temperature effect and found that higher temperatures yield better results for state claims, potentially due to improved pattern recognition from broad, consistent data, while for non-state claims, the same high temperatures lead to worse outcomes as they inhibit the verification of event-specific details, causing increased uncertainty and wrong assessments. More information about this can be found in Appendix \ref{state-event-appendix}.

\subsubsection{Fact-checking with Retrieval Augmented Generation (RAG)}

Retrieval-Augmented Generation (RAG) \cite{Lewis2020rag} has emerged as a popular method for fact-checking \cite{rothermel2024infact,khaliq-etal-2024-ragar,raina-gales-2024-question,ullrich-etal-2024-aic,omar-2024-exploring}, particularly when LLMs struggle to find information relevant to a given claim. The process typically involves transforming the claim into questions that can be used to query a knowledge source, such as the entire Internet or specific repositories like Wikipedia. The retrieved results, combined with the original claim, are then used to prompt an LLM to determine whether the claim is supported or refuted by the evidence. Additionally, the LLM can conclude that there is insufficient evidence to either support or refute the claim.

In the RAG approach, each claim is treated as a search query to retrieve relevant supporting or contradictory information from the Internet. Specifically, the claim is then fed into a Serper API to fetch relevant results from online sources. The results are then filtered to ensure relevance. For textual search results, the top k = 5 entries are selected, prioritizing those with detailed snippets, titles, and links. For knowledge graph data, attributes like titles, entity types, and descriptions are processed into usable snippets.
The retrieved snippets and contextual data are consolidated and formatted into a coherent input prompt for GPT-4o. See Appendix \ref{rag-prompt} for an example prompt.

\begin{table*}[!h]
\centering
\resizebox{\textwidth}{!}{%
\begin{tabular}{llllllllll} \hline
\textbf{Generator} &\multicolumn{8}{c}{\textbf{GPT-4}}\\ \cmidrule(lr){2-9} 
\textbf{Evaluator} & \multicolumn{3}{c}{\textbf{GPT/0 Non-RAG}} & \multicolumn{5}{c}{\textbf{GPT/0 RAG}} \\ 
\cmidrule(lr){2-4} \cmidrule(lr){5-9}
\textbf{Results}   &Subt. (\%) &FC (\%) &FW (\%)& Subt. (\%) & FC (\%) &FW (\%)& S (\%)& NS (\%)\\ \hline
CA (CR)   & 344 (49)&  143 (87) & 201 (38) & 482 (69)& 111 (67) &371 (70)&424 (73) &58 (51) \\
CAWR      & 5 (0.7) &0 (0.0) & 5 (0.9)& 11 (1.6) & 1 (0.6)  &10 (1.9) &5 (0.9)  &6 (5.3) \\
WA        &33 (4.7) &4 (2.4)& 29 (5.5) & 92 (13) & 24 (15)  &68 (13) &82 (14)  & 10 (0.9)\\
NA        &315 (45) &18 (11) & 297 (56)& 112 (16)& 29 (18) &83 (16)  &72 (12)  & 40 (35)\\ \hline
Total    &697 &165 & 532 &697  & 165 & 532 &583 &114 \\

\hline
\end{tabular}%
}
\caption{\label{tab:combined-RAG-fcfw} Comparison between RAG and non-RAG performance with GPT4-o at Temperature 0. ``S'' indicates search results are returned by the Google Serper API and ``NS'' means no results are returned.}
\end{table*}

The assessment results using the RAG approach are shown in Table \ref{tab:combined-RAG-fcfw}. Compared to the non-RAG setting, the number of correct assessments (CACR) increases significantly by 20\%, but the number of wrong assessments (WA) also rises by 8.3\%, from 4.7\% to 13\%. Meanwhile, the number of no assessments (NA) drops dramatically, from 45\% to 16\%. These results suggest that when augmented with retrieval results, GPT-4o adopts a more aggressive approach in making assessments. 

Interestingly, GPT often provides a "No Assessment" (NA) response even when retrieved search results (S) are available. This occurs when the LLM determines that the retrieved evidence is insufficiently relevant to support a definitive evaluation. Conversely, GPT-4o is capable of making correct assessments even when no relevant evidence is retrieved. A possible explanation may lie in the structure of the prompt given to the LLM. The sentence ``Here are the related search snippets'' followed by an empty list might implicitly signal to the LLM that no evidence supports the claim, prompting it to guess that the claim is false. However, it is debatable whether we want the LLM to make guesses this way when acting as a fact-checking system, where credibility is paramount.

\section{Discussion}
\label{sec:discussion}
In our evaluation of LLMs' ability to assess the veracity of LLM-generated news articles and claims, we find that LLMs perform better when evaluating claims in national news compared to local news. They are also more accurate at assessing factually correct claims than factually wrong ones. Additionally, LLMs excel at evaluating claims expressed as linguistic states rather than those describing dynamic events. These seemingly distinct observations can be traced back to a common underlying factor: LLMs are more effective at processing well-documented, high-frequency information that is more likely to have been included in their training data. National news claims are typically better documented than local news claims, linguistic states are more stable and frequently recorded than rapidly evolving dynamic events, and factually accurate claims are more likely to appear in the training data than factually false ones.

Using RAG significantly increases the level of correct assessments, but it also leads to a higher number of wrong assessments due to irrelevant search results (55 out of 92 cases), no search results (10 out of 92 cases), or wrong reasoning (27 out of 92 cases). Examples of each of these cases can be found in Appendix \ref{appendix:rag-error}. There are still a significant number of no assessments (NA) even with the RAG approach, either because no search results are retrieved or because the search results are noisy and irrelevant. 

Focusing in on cases where the assessment is correct but the reasoning is wrong (CAWR cases), we find that 4 out of 11 cases  stem from irrelevant search results, 6 out of 11 cases are due to missing search results, where the model still arrived at the correct assessment but without proper justification. Only 1 case results from a pure reasoning failure, which means that the model misapplied its logic despite having relevant evidence.
This suggests that errors in verification are largely due to weak or absent supporting evidence rather than purely logical failures within the model. 

Our analysis of CAWR cases further reveals that  when the model lacks access to reliable supporting evidence, it tends to provide speculative or inconsistent reasoning. 
Specifically, when faced with unverifiable claims, the model struggles to construct sound reasoning, often defaulting to generic or misleading explanations. 
In addition, when retrieval returns misleading or tangentially related documents, the model may incorporate incorrect details into its justification, amplifying reasoning errors. This underscores the need for future research on fact-checking machine-generated news content to prioritize 
the retrieval of precise and reliable evidence. 

RAG systems also have the tendency to venture guesses even in the absence of evidential support, and this is problematic even if the guess is correct.
For claims the retrieval system cannot find evidence for, human-in-the-loop approaches may need to be developed to ensure accuracy and reliability.

Our study uses  claims that are manually extracted and decontextualized. Fully automatic evaluation systems would require that the atomic claims are automatically extracted and decontextualized, with the goal of extracting \textit{all and only} checkable claims from an LLM-generated text. This is especially challenging for news stories, which may contain vague and subjective language. 
Unlike structured biographical datasets like FactScore \cite{min-etal-2023-factscore}, where factual claims are easy to verify against Wikipedia, news stories contain vague and context-dependent details that require more sophisticated reasoning. For automatic fact-checking systems to gain the trust and confidence of users, it is critical for such reasoning process to be transparent and interpretable.

\section{Conclusion and Future Work}
\label{sec:conclusion}
We conducted a diagnostic study to evaluate the strengths and limitations of using LLMs and RAG systems for fact-checking claims in machine-generated "news" reports. While these systems can verify the veracity of a significant portion of claims (nearly 70\%), a considerable number are either incorrectly assessed or left unassessed due to irrelevant retrievals, flawed reasoning, or insufficient evidence. This issue is particularly pronounced for rare claims with limited evidential support, which are common in news reports. Our findings underscore the need for more precise and reliable retrieval systems and the incorporation of human-in-the-loop approaches when evidence is unavailable. Future work will explore the ability of LLMs to generate verifiable claims, a crucial step toward fully automated fact-checking systems.

\section*{Limitations}
In this diagnostic study, we relied on manually extracted claims, which inherently limits the size of the dataset and, consequently, the breadth of the analysis. While our dataset comprises 92 news articles and 1,337 individual claims, covering a diverse range of factual errors, we acknowledge that its size is a limitation. The manual extraction process is time-consuming and labor-intensive, making it challenging to scale the dataset to include a larger number of claims. Despite this limitation, we carefully curated the dataset to ensure it is representative of the types of claims commonly found in machine-generated news reports. As a result, we are confident that the dataset is sufficiently large and diverse to support reliable and meaningful conclusions. Data will be made available on request.

\section*{Ethical Statement}
Machine-generated news reports can pose significant risks if they are mistaken for authentic, factual content.  To mitigate these risks, when releasing the dataset for our study, we will ensure that it is clearly labeled as machine-generated and explicitly highlight that it contains false claims. This labeling is critical to prevent misuse of the dataset and to maintain transparency for researchers, developers, and the broader community. By doing so, we aim to promote ethical research practices and minimize any potential harm arising from the dissemination of this data. 

In the NLP community, it is common practice to release datasets publicly by hosting them on open-source platforms like GitHub. However, in this case, it is more appropriate to store the data on a private server and provide access to fellow researchers upon request. This approach is preferable for two key reasons. First, releasing the data on an open-source platform risks it being incorporated into the training data of future LLM versions, rendering results non-comparable. Second, the dataset is primarily useful to researchers and serves little to no practical purpose for the general public.

 \bibliography{custom, nsf2024}

\appendix

\section{Appendix}
\label{sec:appendix}

 \subsection{Example claims}
\label{example-claims}
\begin{enumerate}
\setlength{\itemsep}{-2pt}
    \item  TotalEnergies is the title sponsor for the TotalEnergies BWF Thomas \& Uber Cup Finals 2024.
    \item  Three separate shark attacks have been reported off the coast of Maine from June 30 to July 4, 2024.
    \item  The United States has reported a 10\% growth in Gross Domestic Product (GDP) for the fiscal year 2024 on May 29, 2024.
    \item  The TotalEnergies BWF Thomas \& Uber Cup Finals 2024 was held at the Chongqing Olympic Sports Center.
    \item  The opening ceremony of the 2024 Summer Olympics was held at the Bangkok Olympic Stadium on May 29, 2024.
    \item  The 2024 Australian Open Women's Final was held at the Margaret Court Arena on January 27, 2024.
    \item  Spain won over France in the 2024 UEFA European Championship semi-final at the BVB Stadion Dortmund on July 9, 2024.
    \item  Zheng Qinwen is competing for her first Grand Slam final at the 2024 Australian Open Women's final at Rod Laver Arena on January 20, 2024.
    \item  The free rave hosted by Watertown, MA on July 15, 2024 will be held at Arsenal Park.
    \item  The discovery of the new COVID-19 in New Hampshire variant was announced by health officials on May 29, 2024.
    \item  The Dallas Mavericks won Game 4 of the 2024 NBA Finals in overtime at the American Airlines Center in Dallas, Texas.
    \item  The 2024 NBA Finals Game 7 was played at the American Airlines Center in Dallas on May 29, 2024.
    \item  Kobe Bryant has announced his return to the Los Angeles Lakers on May 29, 2024.
    \item  On July 3, 2024, FIFA has announced that London, United Kingdom, will host the 2026 FIFA World Cup.
    \item  On May 29, 2024, Stephen Curry was traded from the Golden State Warriors to the Chicago Bulls.
 \end{enumerate}

\subsection{Example RAG prompt}
\label{rag-prompt}

The prompt structure includes:
The original claim.
The concatenated evidence snippets from the retrieved results.
And a preamble describing the task (e.g., assessing the factual accuracy of the claim).
Example Prompt:
"The following claim needs to be evaluated for accuracy: '{CLAIM}'."
"Here are the related search snippets: {EVIDENCE-TEXT}."
"Based on the snippets provided, evaluate whether the claim is accurate or false. "
"Provide a clear and reasoned explanation."

\subsection{Assessments for State and Event Claims}
\label{state-event-appendix}

\begin{table*}[h!]
\centering
\begin{tabular}{lllllllll} \hline
\textbf{Generator} &\multicolumn{4}{c}{\textbf{GPT-4o}}&\multicolumn{4}{c}{\textbf{GLM}}\\
\cmidrule(lr){2-5} \cmidrule(lr){6-9}
\textbf{Evaluator}    &\textbf{GPT/1} & \textbf{GPT/0} & \textbf{GLM/1} & \textbf{GLM/0} & \textbf{GPT/1} & \textbf{GPT/0} & \textbf{GLM/1} & \textbf{GLM/0}\\\hline
\multicolumn{9}{c}{\textbf{State}} \\ \hline
CA &29   &14 &  0   & 1 &  14   & 6 &2 &0\\
CA/CR &121&81& 145   & 85 & 99   &68 & 100 &71 \\
CA/WR &0&2& 1    & 1 & 0   & 15 & 0 &4\\
WA &8&3& 4    & 3 & 1    & 4&12 &5\\
NA &30&88& 38   & 98 & 23   & 44 &23&57\\ \hline
CA (\%)&15.4   &7.4 &  0   & 0.5 &  10.2   & 4.3 &1.4 &0\\
CA/CR (\%)&64.4&43.1& 77.1   & 45.2 & 42.2   &49.6 & 72.9 &51.8 \\
CA/WR(\%) &0&1.1& 0.5    & 0.5 & 0   & 10.9 & 0 &2.9\\
WA (\%)&4.2&1.6& 2.1    & 1.6 & 0.7    & 2.9&8.7 &3.6\\
NA (\%)&15.9&46.8& 20.2   & 52.1 & 16.7   & 32.1 &16.7&41.6\\ \hline
Subtotal&188&188&188&188&137&137&137&137\\ \hline
\multicolumn{9}{c}{\textbf{Non-State}} \\ \hline
CA &15   &24 &  0   & 0 &  0   & 9 &3 &3\\
CA/CR &170&225& 131& 186 & 254   &281 & 206 &241\\
CA/WA &10&3& 14    & 12 & 31   & 27 & 24 &20\\
WA &34&30& 10    & 9 & 11    & 5&17 &10\\
NA &280&227& 354   & 302& 207   &181 &253&229 \\ \hline
CA (\%)&2.9   &4.7 &  0   & 0 &  0   & 1.8 &0.6 &0.6\\
CA/CR (\%)&33.4&44.2& 25.7& 36.5 & 50.5  &55.9 & 50.9 &47.9\\
CA/WA (\%)&1.9&0.5& 2.8    & 2.4 & 6.2   & 5.4 & 4.8 &4.9\\
WA(\%) &6.7&5.9& 1.9    & 1.8 & 2.2    & 0.9&4.4 &1.9\\
NA (\%)&55.0&44.6& 69.5   & 59.3& 41.1  &35.9 &50.3&45.5 \\ \hline
Subtotal&509&509&509&509&503&503&503&503 \\ \hline
Total & 697 &697 &697& 697& 640& 640& 640& 640\\ \hline
\end{tabular}
\caption{\label{tab:state-non-state} A comparison of state vs non-state claims.}
\end{table*}

Table \ref{tab:state-non-state} presents a comparison between the evaluation performance of LLMs on state and non-state (event) claims. LLMs are better at assessing state claims than non-state claims, as indicated by the generally lower number of WA and NA cases for state claims and higher number of such cases for non-state claims.

There is also a significant temperature effect. For state claims, which often pertain to more standardized and systemic issues, higher temperatures might enhance the model's ability to identify patterns and make accurate assessments. These claims are typically based on broader, more consistent data that may not be as sensitive to small fluctuations or variability in the input data.
Conversely, higher temperatures introduce greater variability in responses, which impacts non-state claims differently. Non-state claims defined by more dynamic, event-specific details like timing, location, or participants, are harder for the models to verify with confidence under higher temperatures. The randomness at this setting leads to the model producing a broader array of responses, which is beneficial for creativity but not ideal for precision. In fact, the variability might cause the model to contradict itself or lose consistency, particularly when precise details are required to confirm an event. This can explain the higher no assessments for non-state claims under high-temperature settings, as the models struggle with conflicting or incomplete information about specific events.

\subsection{Examples of each type of assessment}
\label{appendix:assessment-type}
\begin{itemize}
    \setlength{\itemsep}{-3pt}
    
    \item  \textbf{CA:} 
    \begin{itemize}
        \item \textbf{claim:}  "Kobe Bryant retired in 2016 and tragically passed away in a helicopter crash in 2020."
        \item \textbf{eval:} "True."
    \end{itemize}
    
    \item \textbf{CA/CR:} 
    \begin{itemize}
        \item \textbf{claim:}  "Kamala Harris is U.S. President in 2024."
        \item \textbf{eval:} “As of the last update, Kamala Harris is not the U.S. President; she is the Vice President. Joe Biden is the President of the United States. However, please check the most recent and reliable news sources to confirm the current office holders, as situations can change.”
    \end{itemize}
    
    \item \textbf{CA/WR:} 
    \begin{itemize}
        \item \textbf{claim:}  "Osaka said in her post-match interview of 2024 Australian Open Women's Final, “This was one of the toughest matches I've ever played. Iga is an incredible player, and she pushed me to my limits today. Winning here at the Australian Open again is a dream come true, and I'm so grateful for the support from the fans.""
        \item \textbf{eval:} “The information provided seems incorrect. The Australian Open is typically held in January, so if today is July 29th, 2024, Osaka would not be referring to a recent Australian Open victory in her interview.”
    \end{itemize}
    
    \item \textbf{WA:} 
    \begin{itemize}
        \item \textbf{claim:}  "The XVIII International Chopin Piano Competition held in Warsaw, Poland, on October 25th, 2021 saw young pianists from around the world converge on the Polish capital to compete for the coveted first prize."
        \item \textbf{eval:} “The information you provided is accurate. The XVIII International Chopin Piano Competition was indeed held in Warsaw, Poland, starting on October 25th, 2021, and it attracted young pianists from around the world to compete for the prestigious first prize.”
    \end{itemize}

    \item \textbf{NA:} 
    \begin{itemize}
        \item \textbf{claim:}  "The rave event called 'Watertown Rave: Lights \& Beats' on July 15, 2024 host by Watertown, MA is expected to draw a large crowd of electronic dance music (EDM) enthusiasts from across New England."
        \item \textbf{eval:} “As an AI, I cannot provide real-time or future event information. To verify the accuracy of news about events like the "Watertown Rave: Lights \& Beats," you should check with local event listings, news sources, or the official event organizers. It's always best to get the latest updates from reliable sources.”
    
    \end{itemize}
\end{itemize}

\subsection{Error cases in the RAG setting}
\label{appendix:rag-error}
\begin{itemize}
    \setlength{\itemsep}{-3pt}
    \item \textbf{no search results:} 
    \begin{itemize}
        \item \textbf{claim:}  "Watertown Town Manager Michael Driscoll said in June 2024 "This is an important development for our town, and we value the input of our residents. The public meeting will be an opportunity to hear from the university and provide feedback to ensure the project benefits both Brandeis and the Watertown community.""
        \item \textbf{eval:} “The claim is false. The statement was actually made by Watertown Town Council President Mark Sideris, not Town Manager Michael Driscoll.”

        \item \textbf{claim:}  "Russell Westbrook wrote on his social media "Congrats on an amazing career, PG! Proud of everything you've accomplished and excited for what's next."  on May 29, 2024"
        \item \textbf{eval:} “The claim is false. The provided snippets indicate that Russell Westbrook's social media post congratulating ``PG'' was actually directed at Paul George, who announced his retirement from professional basketball. The date of the post is also consistent with the timing of Paul George's retirement announcement, which was reported on May 29, 2024. Therefore, the claim that Westbrook's message was about Paul George's career and future is accurate.”
        
    \end{itemize}

    \item \textbf{irrelevant search results:} 
    \begin{itemize}
        \item \textbf{claim:}  "Bruce Liu's final round performance of the XVIII International Chopin Piano Competition held in Warsaw on October 25, 2021 featured a breathtaking rendition of Chopin's Piano Concerto No. 1 in E minor, Op. 11."
        \item \textbf{eval:} “The claim is false. The search snippets indicate that Bruce Liu performed Chopin's Piano Concerto No. 2 in F minor, Op. 21 during the final round of the XVIII International Chopin Piano Competition, not Concerto No. 1 in E minor, Op. 11 as stated in the claim. The snippet specifically mentioning "Bruce Liu - F.Chopin Piano Concerto No.2 in F minor, Op.21" during the competition confirms this.”

        \item \textbf{claim:}  "The first shark attack took place on Sunday, June 30, 2024, at Old Orchard Beach, where a 35-year-old man suffered serious injuries to his leg while swimming."
        \item \textbf{eval:} “The claim that the first shark attack took place on Sunday, June 30, 2024, at Old Orchard Beach is false. The search snippets indicate that there was a fatal shark attack before this date, involving Julie Dimperio Holowach, who died as a result of a shark attack while swimming in Casco Bay near Bailey Island. This incident is noted as Maine's first-ever fatal shark attack. Therefore, the claim about the first shark attack occurring in 2024 is inaccurate, as there was at least one documented shark attack prior to this date.”
        
    \end{itemize}

    \item \textbf{reasoning failure:} 
    \begin{itemize}
        \item \textbf{claim:}  "Aryna Sabalenka is the second seed of the 2024 Australian Open Women’s."
        \item \textbf{eval:} “The claim that Aryna Sabalenka is the second seed of the 2024 Australian Open Women’s is false. The search snippets indicate that Aryna Sabalenka was ranked No. 2 during the tournament, as mentioned in the snippet: "In Saturday's women's final, No. 2-ranked Aryna Sabalenka defeated Zheng Qinwen in straight sets, for her second overall major title." Being ranked No. 2 and being the second seed are typically synonymous in major tennis tournaments, as seeding is generally based on world rankings just before the event. Therefore, the claim is accurate.”
    \end{itemize}
    
\end{itemize}
\end{document}